\documentclass[conference]{IEEEtran}
\IEEEoverridecommandlockouts
\usepackage{amssymb}
\usepackage{latexsym}
\usepackage{caption}
\usepackage{verbatim}
\usepackage{algorithm}
\usepackage{algorithmic}
\usepackage{tabularx}
\usepackage{amssymb}
\usepackage{amsmath}
\usepackage{amssymb}
\usepackage{latexsym}
\usepackage{caption}
\usepackage{subcaption}
\usepackage{verbatim}
\usepackage{algorithm}
\usepackage{algorithmic}
\usepackage{tabularx}
\usepackage{amssymb}
\usepackage{amsmath}
\usepackage{mathtools}
\usepackage{color}
\usepackage{multirow}
\usepackage{booktabs}

\def\BibTeX{{\rm B\kern-.05em{\sc i\kern-.025em b}\kern-.08em
    T\kern-.1667em\lower.7ex\hbox{E}\kern-.125emX}}

\newtheorem{theorem}{Theorem}

\begin{document}

\title{PowerLinear Activation Functions with application to the first layer of CNNs\\
{\footnotesize \textsuperscript{*}}
}

\author{\IEEEauthorblockN{1\textsuperscript{st}  Kamyar Nasiri}
\IEEEauthorblockA{\textit{Computer Engineering Department} \\
\textit{Ferdowsi University of Mashhad}\\
Mashhad, Iran \\
kamyarnasiri2000@mail.um.ac.ir}
\and
\IEEEauthorblockN{2\textsuperscript{nd} Kamaledin Ghiasi-Shirazi}
\IEEEauthorblockA{\textit{Computer Engineering Department} \\
\textit{Ferdowsi University of Mashhad}\\
Mashhad, Iran \\
k.ghiasi@um.ac.ir}
}

\maketitle

\begin{abstract}
Convolutional neural networks (CNNs) have become the state-of-the-art tool for dealing with unsolved problems in computer vision and image processing. Since the convolution operator is a linear operator, several generalizations have been proposed to improve the performance of CNNs. 
One way to increase the capability of the convolution operator is by applying activation functions on the inner product operator. In this paper, we will introduce PowerLinear activation functions, which are based on the polynomial kernel generalization of the convolution operator. 
EvenPowLin functions are the main branch of the PowerLinear activation functions. This class of activation functions is saturated neither in the positive input region nor in the negative one. Also, the negative inputs are activated with the same magnitude as the positive inputs.
These features made the EvenPowLin activation functions able to be utilized in the first layer of CNN architectures and learn complex features of input images. 
Additionally, EvenPowLin activation functions are used in CNN models to classify the inversion of grayscale images as accurately as the original grayscale images, which is significantly better than commonly used activation functions.
\end{abstract}


\section{Introduction}
Deep neural networks, in particular, Convolutional Neural Networks (CNNs) \cite{lecun1989backpropagation}, have had a crucial role in the performance improvement of computer vision and image processing fields in the last decade. CNNs are a class of deep neural networks that process imagery data and are being widely used both in research and industry. CNN networks utilize the convolutional layer as the main operator, which mimics primary visual cortex cell operation. 

Traditionally, a convolution layer in CNN uses the inner product operator as its main computational core to measure the similarity between input image patches and some filters (kernels). However, the inner product is a linear operator and it is not capable of learning complex features. One way to remedy this problem is accompanying convolutional layers with nonlinear activation functions. Another remedy is to replace the ordinary convolution operator with more complex operators, such as similarity measure functions.
As an example, Lin et al. introduced "Network In Network"(NIN) in \cite{lin2014network}, which replaced the inner product operator in convolutional layers with multilayer perceptron (MLP). In the NIN proposed architecture, a multiple fully connected layer followed by non-linear activation functions is located for each patch in the receptive field of input images.

In 2016, Cohen et al. introduced SimNets architecture that replaced the inner product operator in the convolutional layers with an inner product in feature space \cite{Cohen2016SimNet}. The feature space is essentially controlled by special kernel functions. SimNet architecture was introduced based on two fundamental operators. The first operator is a substitute for the inner product operator, which is a weighted similarity function such as "\(lp\)" form functions. The next operator is a replacement of non-linear activation functions and pooling layers. The second operator is defined based on the MEX(\(log-mean-exp\)) operator. Cohen et al. proved that the SimNet MLP architecture is associated with kernel machines.

\begin{figure*}[ht!]
	\centering
            \includegraphics[width=.5\textwidth]{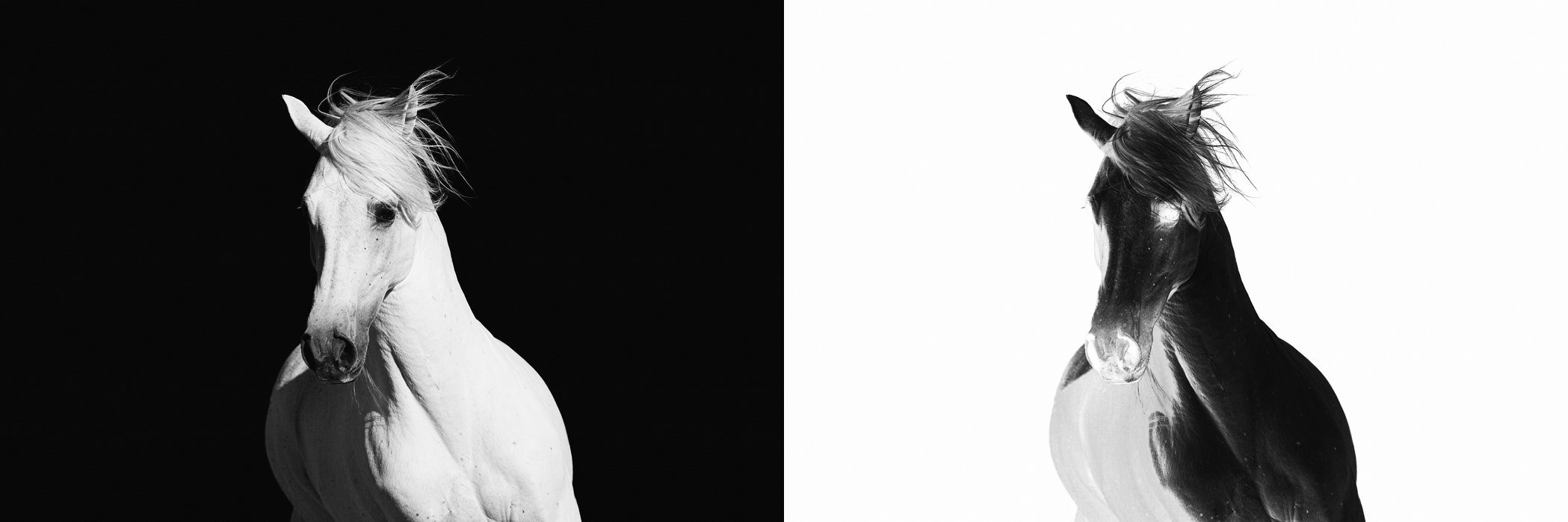}\hfill
            \caption{Example of grayscale image in left, and inversion of that image in right.}
	\label{fig:Horses}
        \end{figure*}

Kernel Pooling was introduced to capture higher-order, non-linear interactions of features in the kernel form \cite{Yin2017KernelPooling}. This method was proposed without additional parameters. It uses Count Sketch to generate a feature map up to \(p\) order. Then, the inner product between two features can capture high order feature interactions by using kernel pooling. In this case, the Discrete Fourier Transforms method was applied \(p\) times. However, higher-order features should be calculated explicitly, which makes the order grow.

Wang et al. introduced Kervolutional Neural Networks(KNNs) as a generalization of CNNs using the kernel methods \cite{Wang2019KNN}. In the architecture of KNN, kernel functions are utilized instead of the inner product operation.
For instance, Polynomial Kervolution was defined in line with the polynomial kernel function.

Basically, since the convolution operator is a linear operator, one way to increase the capability of models and learn more complex features of input images is by applying a non-linear activation function on top of the convolution operator. Thus, non-linearity layers can be applied to convolution layers to obtain non-linear feature maps.
Currently, activation functions are being used as an inseparable component of neural network architectures. Traditionally, Sigmoid and TanH functions were popular activation functions and were used widely in neural network models. The output of the Sigmoid activation function, also called Logistic Sigmoid, is between \(0\) and \(1\). Also, the output of TanH activation function is between \(-1\) and \(1\). These two functions were de facto activation functions for several decades. However, both Sigmoid and TanH functions have the saturation problem. The saturation of activation functions causes the vanishing gradient problem to happen. This problem made those functions unusable for deep neural networks, i.e., neural networks with more than a few layers. 

Rectified Linear Unit (ReLU) solved the saturation problem in the positive input region of activation functions \cite{nair2010rectified}.
Introducing this activation function paved the way for deepening the neural network models and adding more layers to networks. Besides, in this function, the saturation problem in the positive input region is eliminated. The output of the ReLU activation function is linear in the positive inputs region and the constant value of zero in the negative inputs region. 
It should be mention that the saturation problem remains in the negative inputs region of the ReLU activation function. Thus, to solve these problems, several generalizations have been introduced. As an example, the Leaky ReLU activation function \cite{Maas13rectifiernonlinearities} had been proposed to prevent the saturation problem in the negative inputs region by a slight slope for inputs with negative values.

In this paper, we introduce a class of activation functions that have non-zero gradient values in both the positive input and the negative input regions. This class of activation functions, which we call PowerLinear activation functions, does not saturate in any region. Theoretically, we propose this class of activation functions based on the generalized convolution operators introduced in \cite{GhiasiShirazi2019GCNN}. Here, we use the polynomial kernel function for generalizing the dot-product operator in convolution layers. The main type of PowerLinear activation function is called the EvenPowLin activation function, which contains major discussions of this paper. According to our experiments, this function can be applied to the first convolution layer of each CNN. This capability makes some attractive features for convolutional networks.

The interesting behaviour about EvenPowLin activation functions is that they are even functions and produce positive outputs for negative input values. This feature of our proposed functions makes it remarkable among other proposed activation functions.
On the other hand, OddPowLin activation functions, another type of PowerLinear activation functions, are odd functions that activate the negative input values with negative values.
The proposed activation functions are not saturated in the negative input region. 

Additionally, EvenPowLin activation functions provide practical usages. We found an application for EvenPowLin activation functions in making networks that are blind to the colour-inversion of images. For instance, consider a white horse picture on a black background that has been given as a training data image to a network that utilizes commonly used activation functions. The trained model cannot easily diagnose the inversion of the original image (i.e. a black horse image on a white background) as shown in Figure~\ref{fig:Horses}.  Nonetheless, a network equipped with an EvenPowLin activation function can recognize the inverted image as accurately as the original image. 

The paper proceeds as follows. We give some background material in Section~\ref{sec:bg}. Next, we introduce the proposed method in Section~\ref{sec:proposed-method}. In Section~\ref{sec:experiments}, we report the results of our experiments for evaluating the PowerLinear activation functions. We conclude the paper in Section~\ref{sec:conclusion}.

\begin{figure*} [htbp]
\centering
\begin{subfigure}[t]{\textwidth}
\centering
\includegraphics[width=0.5\textwidth]{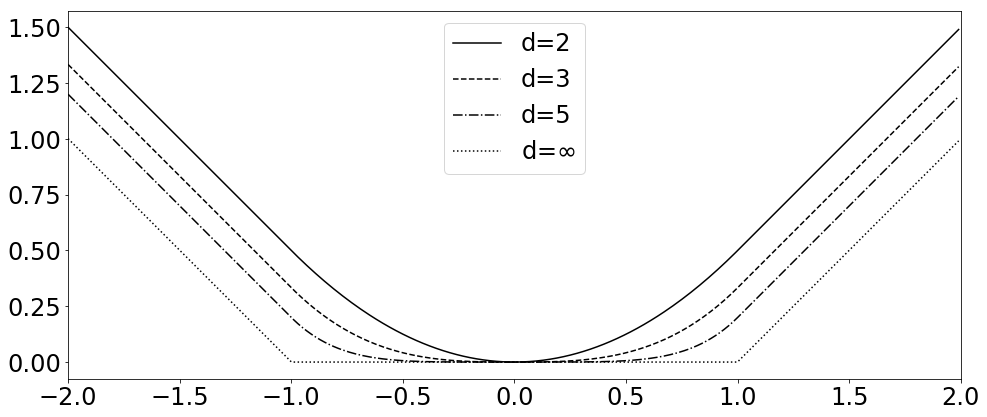}
\caption{}
\end{subfigure}

\hfill
\begin{subfigure}[t]{\textwidth}
\centering
\includegraphics[width=0.5\textwidth]{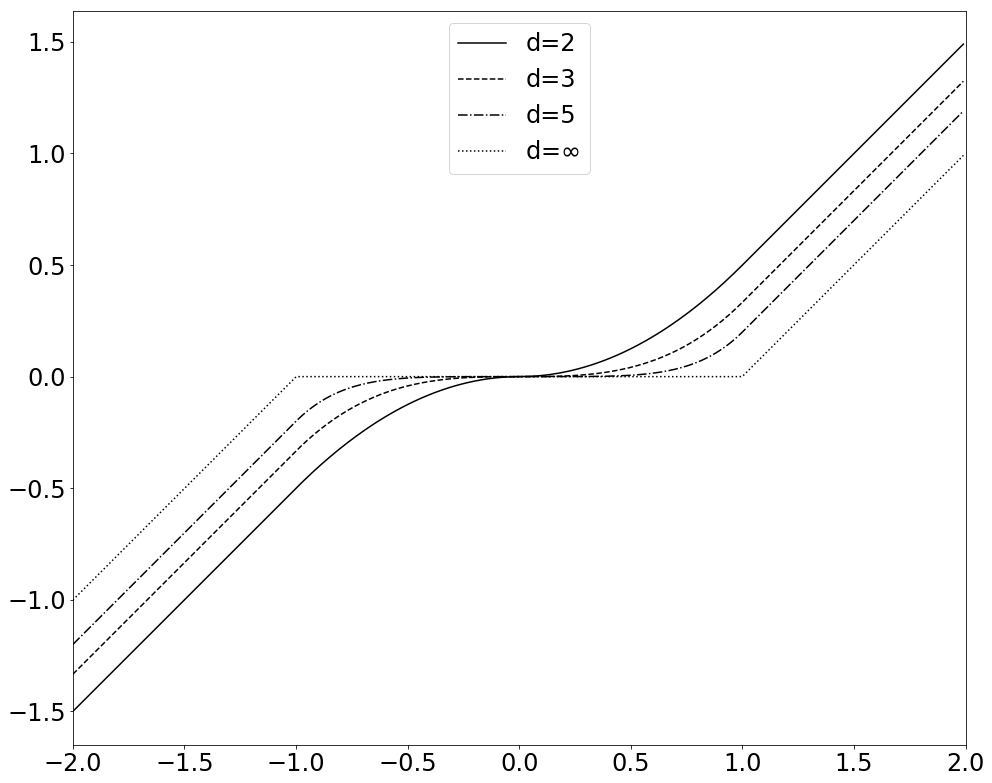}
\caption{}
\end{subfigure}

\caption{PowerLinear activation functions with degrees $d=2,3, 5$, and $\infty$. (a) Even extension, (b) Odd extension.}
\label{fig:PowLin}
\end{figure*}

\section{Background}\label{sec:bg}
As mentioned before, the convolution operator is a linear operation. Two generalizations of the convolution operator were introduced by \cite{GhiasiShirazi2019GCNN} to increase the capability of CNN architectures to learn more complex features.  
The first generalization was defined as followed:

\begin{theorem}[Generalization 1] 
\label{Generalization1}
The convolution operator in CNNs can be generalized by substituting the inner product operation $x^T w$ between a vectorized input $x$ and a weight vector $w$ by $f(k(x,w))$, where $k$ is a positive definite kernel function and $f$ is a monotonically increasing function with $f(0)\geq 0$.
\end{theorem}

According to this generalization, monotonically increasing functions of kernel functions can be utilized inside the convolutional layer.
Additionally, as far as the inner product operator is a similarity measurement function, it is possible to utilize distance measurement functions instead of the inner product.
Therefore, the second proposed generalization of the convolution operator was defined as followed:

\begin{theorem}[Generalization 2]
\label{Generalization2}
The convolution operator in CNNs can be generalized by substituting the inner product operation $x^T w$ between a vectorzied input $x$ and a weight vector $w$ by $f(-d(x,w))$, where $d$ is a distance metric and $f:(-\infty,0]\to [0,\infty)$ is a monotonically increasing function with $f(-\infty)=0$.  Adding the additional constraint $f(0)=1$ ensures that $f(-d(x,x))=1$, meaning that the similarity of each vector $x$ with itself is 1. 
\end{theorem}

Usual non-linear activation functions can be regarded as monotonically increasing functions that act on more basic forms of generalized convolution operators, e.g. Euclidean-distance based convolution operator.

\section{PowerLinear Activation Function}
\label{sec:proposed-method}
Polynomial kernels, along with Gaussian kernels, are the two types of kernels that were most widely used in kernel methods from its advent.
Specifically, polynomial kernels were the default choice of the kernel for digit recognition tasks \cite{scholkopf2002learning}. 
Polynomial kernels are defined as

\begin{equation}
 k(x,w)= (x^Tw+b)^d,  
\end{equation}
where $d\in \mathbb{N}$ is the degree of the kernel.

Theorem~\ref{Generalization1} 
introduces a generalization of the convolution operator using positive definite kernel functions. In this section, we will introduce a generalization of the convolution operator using polynomial kernels, which we call polynomial-measure convolution. Basically, polynomial-measure convolution can be implemented by applying the power activation function $\sigma (x) = x ^ d$ on the output of an ordinary convolution layer.
For odd $d$ the power function $\sigma (x) = x ^ d$ is an odd function and for even $d$ it is an even function.
Note that the odd extension is a monotonically increasing function and therefore a valid activation function according to the framework introduced in \cite{GhiasiShirazi2019GCNN}.
Additionally, the even extension of the power function can be written as:

\begin{equation}
|\langle x, w \rangle|^ d= \left(\langle x, w \rangle^2\right) ^{d/2}
\end{equation}

which is equal to the composition of the polynomial kernel $\langle x, w \rangle^2$ and the monotonically increasing function $\sigma:\mathbb{R}^+\to\mathbb{R}$ defined by the formula $\sigma(x)=x^{d/2}$. 

\begin{table*}\centering
\begin{tabular}{ccc}
\toprule
Dataset & ReLU activation function & EvenPowLin activation function  \\
\midrule
Mnist & 0.9956 & 0.9952 \\
\addlinespace 
SVHN & 0.9555 & 0.9571 \\
\addlinespace 
CIFAR10 & 0.8507 & 0.8447 \\
\bottomrule
\end{tabular} 
\caption{This table shows the accuracy of classification normal test data in three different datasets.}\label{TableNormal}
\end{table*}

One problem with the functionality of even/odd-extended power activation functions is that the gradient explodes outside the range $[-1,1]$.
Even if we use an initialization algorithm that ensures that the input to the power activation function is in the range $[-1,1]$ at the start of the training, this property may be violated as the training proceeds.
We use two mechanisms to solve this problem. First, we normalize the input data in our experiments. Second, we replace the power function with linear functions outside the range $[-1,1]$ to avoid exploding gradients. 
The result of the above line of reasoning is two novel classes of activation functions which we call the EvenPowLin and OddPowLin activation functions - PowerLinear activation functions in general.

The EvenPowLin activation function of order $d$ is defined as:

\begin{eqnarray*} 
\sigma(x) =   \left\{ \begin{array}{ll}
			x-(1-\frac{1}{d})\;\;\;\;& if\;x\ge 1 \\
			\frac{1}{d}|x|^d\;\;\;\;\;\;\;\;& if\;-1\le x < 1 \\						
			-x-(1-\frac{1}{d})\;\;\;\; &if\;x< -1
			
			 \end{array} \right.
\end{eqnarray*}

Similarly, the OddPowLin activation function of order $d$ is defined as:

\begin{eqnarray*} 
\sigma(x) =   \left\{ \begin{array}{ll}
			x-(1-\frac{1}{d})\;\;\;\;& if\;x\ge 1 \\
			\frac{1}{d}|x|^d\;\;\;\;\;\;\;\;& if\;0\le x < 1 \\						
			-\frac{1}{d}|x|^d\;\;\;\;\;\;\;\;& if\;-1\le x< 0 \\			
			x+(1-\frac{1}{d})\;\;\;\; &if\;x< -1
			
			 \end{array} \right.
\end{eqnarray*}

Figure~\ref{fig:PowLin} (a) shows the graph of EvenPowLin activation function and Figure~\ref{fig:PowLin} (b) shows the graph of OddPowLin activation function, both for $d=2,3,5,\infty$.

When an input image patch $x$ is not aligned with a weight vector $w$, i.e. the magnitude of $\langle x, w \rangle$ is small, then both OddPowLin and EvenPowLin activation functions strongly attenuate the dot-product similarity.   
When $x$ and $w$ have a strong positive correlation, i.e. the magnitude of $\langle x, w \rangle$ is large, then both OddPowLin and EvenPowLin produce positive similarity values.
The interesting case is when $x$ and $w$ have a strong negative correlation. In this case, the OddPowLin activation function produces a strong negative value, indicating strong dissimilarity, while the EvenPowLin activation function outputs a strong positive value, accepting both the similarity to $w$ and $-w$.
This analysis provides a new explanation about why polynomial kernels have performed well in kernel methods and justify the usage of even activation functions.

PowerLinear activation functions are zero-centered and differentiable at all points.
PowerLinear activation functions do not eliminate the activity of any neurons, i.e. direct the activity of neurons with negative outputs toward non-zero values- positive values for EvenPowLin functions and negative values for OddPowLin functions. 

In the following section, we will discuss the experiments on the EvenPowLin activation functions as the activation function applied to the first layer of CNN architectures. Basically, our main discovery in this paper belongs to the EvenPowLin functions. It is because EvenPowLin functions are even functions and activate negative inputs as well. This attribute is unique among other activation functions, which can lead us to practical applications that will be discussed. 

\section{Experiments}
\label{sec:experiments}
In this section, we will demonstrate some possible applications of EvenPowLin activation functions. We should emphasize that EvenPowLin activation functions will be applied on just the first layer of CNN architectures during the experiments. 

First of all, we found that the performance of EvenPowLin activation functions is similar to the performance of the ReLU activation function in terms of classifying normal datasets. Normal datasets
are those which are used preparedly and are not inverted, such as Mnist and SVHN datasets. 

The second category of our experiments refers to the special functionality that we expected to get better results by applying EvenPowLin activation functions to convolution layers. This functionality is the fact that our proposed functions should make the neural networks persistent against the inversion of grayscale images. In this case, we inverted some grayscale images based on normal image datasets. Fortunately, we noticed a significantly better result than applying other activation functions instead. 

We will discuss these two major parts of our experiments in the following. 

At first, we examined the general functionality of our proposed activation functions. Basically, our proposed class of activation functions can be used in rudimentary CNN architectures as the activation function applied to the first convolution layer. 

As the first case of the experiments, we implemented ResNet-18 architecture. This architecture is a modern CNN architecture and it has been widely used since its creation. ResNet-18 is a convolutional neural network that is considered a deep neural network. We utilized this architecture to classify the Mnist dataset images, SVHN dataset images, and also CIFAR-10 dataset images.

\begin{table*}[ht!]\centering
\begin{tabular}{ccc}
\toprule
Dataset &  Normal data (ReLU) & Inverted data (ReLU)  \\
\midrule
Mnist & 0.9952 & 0.1136 \\
\addlinespace 
SVHN &  0.9451 & 0.9292 \\
\bottomrule
\end{tabular}
\caption{This table shows the accuracy of classification normal test data and inversion of test data using ReLU activation function in two different datasets.}\label{ReluInvert}
\end{table*}

\begin{table*}\centering
\begin{tabular}{ccc}
\toprule
Dataset & Normal data (EvenPowLin) & Inverted data (EvenPowLin)  \\
\midrule
Mnist & 0.9948 & 0.9948\\
\addlinespace 
SVHN & 0.9412 & 0.9412 \\
\bottomrule
\end{tabular}
\caption{This table shows the accuracy of classification normal test data and inversion of test data using EvenPowLin activation function in two different datasets.}\label{PowlinInvert}
\end{table*}

We implemented ResNet-18 architecture by applying the ReLU activation function on the output of each convolution operator layer. After that, we loaded the Mnist dataset and normalized it to classify the input images into digit labels between 0 and 9. As shown in Table~\ref{TableNormal}, the accuracy of classification on the test Mnist dataset using ResNet-18 architecture is 99.56\%. 

In the next experiment, we examined the performance of EvenPowLin activation functions in the same condition of the previous experiment. In our experiments, We decided to utilize only the power {2} of EvenPowLin activation functions and apply this function only on the output of the first convolution layer. The result of using the EvenPowLin activation function with power {2} is shown in Table~\ref{TableNormal}. It is obvious that the accuracy of both ReLU and EvenPowLin activation functions are similar to each other in this case. It should be noticed that the number of epochs in these two experiments was 10 with a batch size equal to 256 and the learning rate was set to 0.01.

Similarly, we repeated these experiments for SVHN and CIFAR10 datasets. SVHN dataset is an image dataset including 10 classes. Each class is dedicated to a digit. According to Table~\ref{TableNormal}, the accuracy of classifying the test dataset using EvenPowLin activation functions for SVHN is 0.9571, and for the CIFAR10 is 0.8447. We have set the number of epochs for training both of these datasets to 15. 

In the next experiments, we will discuss the main functionality of the EvenPowLin activation functions. As mentioned before, EvenPowLin activation functions have the ability to activate equal-value positive and negative inputs with the same output values. To show this point, we just examined grayscale images. In this case, if the output of the input neuron is bigger than 1, it will be activated linearly, which is similar to the ReLU activation function. 
Similarly, if the output of the input neuron is less than {-1}, it will be activated linearly with a negative slope. The value of the negative slope is the same as the value of the positive slope in the positive input region. 
Thus, in grayscale images, white pixels can be activated with the same value as black pixels. From the perspective of EvenPowLin activation functions, the darkness of black pixels and the lightness of white pixels are equal.

\begin{figure*}[t!]
    \centering
    \begin{subfigure}[t]{0.3\textwidth}
        \centering
        \includegraphics[height=1.2in]{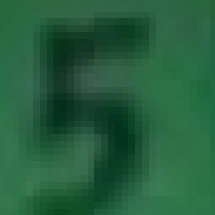}
        \caption{First test image in the SVHN dataset} 
    \end{subfigure}%
   \begin{subfigure}[t]{0.3\textwidth}
        \centering
        \includegraphics[height=1.2in]{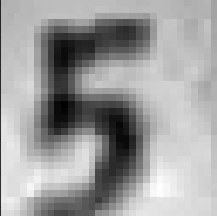}
        \caption{Grayscale conversion of the first test image in the SVHN dataset} \label{Svhn2}
    \end{subfigure}%
    \begin{subfigure}[t]{0.3\textwidth}
        \centering
        \includegraphics[height=1.2in]{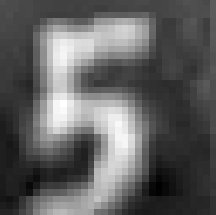}
        \caption{Inversion of grayscale conversion of the first test image in the SVHN dataset} \label{Svhn3}
    \end{subfigure}
    \caption{First test image in SVHN dataset}
\label{fig:Svhn}
\end{figure*}

In the following experiments, we examined our proposed activation functions on two image datasets, the Mnist dataset, and the SVHN dataset. By applying the EvenPowLin activation function on the output of the first convolution layer in a CNN model to classify normal grayscale images into digit labels, the model can diagnosis the inversion of images accordingly. The inversion of a grayscale image is the negation of all its pixel values. For instance, if the Mnist dataset images are white digits on black background, the inversion of these images will be a black digit on white background.

One of our innovations to make a better result was converting the range of input data pixels between {-1} and {1}. It seems necessary since our proposed function is zero-centered. After finishing the training process on the normal Mnist train dataset, we inverted the Mnist test dataset. Then, we tested both the normal Mnist test dataset and the inverted test dataset to show the comparison. Basically, we expected that both have the same accuracy because the dark pixels activate with the same value as the light pixels in the training process.

In the beginning,  we implemented the ResNet-18 architecture by applying the ReLU function over all the convolution layers. The accuracy of testing the model on normal Mnist test data was 99.48 \% and the accuracy of testing the model on inverted Mnist test data was approximately 11.36 \%, as shown in Table~\ref{ReluInvert}.

In the next step, we implemented ResNet-18 architecture again and set the EvenPowLin activation function on the output of the first convolution layer and the ReLU activation function on other layers. Then, we selected normal training data of the Mnist dataset as the training dataset. The accuracy of testing our model on the inverted Mnist test came to 99.48 \%, with the same accuracy as normal test data reached, as shown in Table~\ref{PowlinInvert}. It is almost 88 \% more than the accuracy obtained by the network equipped by only the ReLU activation function.

In the next experiment, we focused on the SVHN dataset. Similar to the Mnist dataset, the task is to classify input digits, which are the house numbers in this case. 
According to the fact that the images in the SVHN dataset are colorful, to obtain better efficiencies, we converted the images in this dataset to grayscale images. Figure~\ref{fig:Svhn} (a) demonstrates the original first test image of the SVHN dataset, Figure~\ref{fig:Svhn} (b) shows the grayscale conversion of the image in Figure~\ref{fig:Svhn} (a), and Figure~\ref{fig:Svhn} (c) is the inversion instant of Figure~\ref{fig:Svhn} (b). 

In this case, it should be mentioned that the grayscale SVHN training dataset has both black digit images on white backgrounds and white digit images on black backgrounds. Naturally, it is because these images were colorful originally, and we converted them to grayscale. 
Hence, it should be noticed that the ReLU activation function will be applied to both kinds of grayscale images during the training process. 
In this case, a model that includes the ReLU activation function on the first convolution layer will diagnose white digits on black backgrounds as well as black digits on white backgrounds and they are somehow persistent against the inversion of grayscale images. However, since the number of black digit images on white backgrounds is much more than white digit images on black backgrounds in the training dataset, the model with only the ReLU activation function on each layer will be dependant on this type of grayscale image.

In the final experiment case, We examined the SVHN dataset using the ResNet-18 architecture equipped with the ReLU activation function on the output of all the convolution layers. In terms of classifying normal grayscale test images of the SVHN dataset, the accuracy reached 94.51 \%. As described before, we expected using the ReLU function on the top of all convolution layers would be persistent against inversion. Thus, the accuracy of classifying the inverted test images peaked at 92.92 \%. Fortunately, with changing the first applied activation function to the EvenPowLin activation function, the accuracy stood at 94.51 \%, which is equal to the accuracy that normal SVHN data reached. This property of EvenPowLin activation functions makes them remarkable and superior to commonly used activation functions in this case.

\section{Conclusions}
\label{sec:conclusion}
In this paper, we proposed a class of activation functions that can be utilized within the Convolutional Neural Networks. PowerLinear activation functions are based on the Polynomial kernel function, which is a generalization of the convolution operator. This class of activation functions prevents the saturation problem in both the positive input region and the negative input region. EvenPowLin activation functions are the main type of PowerLinear activation functions, which are even functions and activate the negative input values with positive values.
We examined EvenPowLin activation functions to classify normal image datasets and inversion of grayscale image datasets. EvenPowLin activation functions perform superior to commonly used activation functions in the case of diagnosing the inversion of grayscale images.

\bibliographystyle{IEEEtran}
\bibliography{IEEEabrv,PowerLinear}

\end{document}